\documentclass[conference]{IEEEtran}
\IEEEoverridecommandlockouts

\usepackage{cite}
\usepackage{amsmath,amssymb,amsfonts}
\usepackage{algorithmic}
\usepackage{graphicx}
\usepackage{textcomp}
\usepackage{xcolor}
\def\BibTeX{{\rm B\kern-.05em{\sc i\kern-.025em b}\kern-.08em
    T\kern-.1667em\lower.7ex\hbox{E}\kern-.125emX}}
\usepackage{booktabs}

\usepackage{comment}

\begin{document}

\title{Measuring Fairness in Large Audio Language Models via Semantic-Aware Bias Estimation}

\author{\IEEEauthorblockN{Zhe Liu}
\IEEEauthorblockA{\textit{Meta Platforms, Inc.} \\
Menlo Park, USA}
}

\maketitle

\begin{abstract}
Large Audio Language Models (LALMs) have seen increasing use for audio understanding tasks such as speech recognition and audio question answering, raising concerns about fairness across demographic subgroups. Fairness evaluation in spoken-input settings is challenging due to confounding factors, including semantic variation in spoken content and speaker-specific characteristics. Ignoring these factors can result in misleading conclusions about model bias. We propose a semantic-aware mixed-effects regression framework for fairness evaluation in LALMs that explicitly accounts for these confounders. Our approach incorporates sentence-level semantic embeddings of reference text as covariates and models speaker identity as a random effect. Notably, semantic representations are extracted from the same LALM under evaluation, enabling semantic control over variation as perceived by the model itself. Experiments on simulated data and real-world benchmarks demonstrate that the proposed approach substantially reduces spurious fairness findings and yields more robust and interpretable estimates of subgroup performance differences.
\end{abstract}


\section{Introduction}
Over the past few years, Large language models (LLMs) have achieved strong performance across a wide range of language understanding and generation tasks \cite{achiam2023gpt, grattafiori2024llama}. More recently, Large Audio Language Models (LALMs), which extend LLMs to process spoken inputs and generate textual or spoken outputs, have emerged as a powerful paradigm for audio understanding and spoken interaction \cite{wu2023decoder, borsos2023audiolm, chu2024qwen2, defossez2024moshi}. These models enable a variety of real-world applications, including automatic speech recognition (ASR), audio question answering, and conversational assistants.

A growing body of studies has shown that LLMs can exhibit biases associated with demographic and social attributes \cite{li2023survey}. When extended to spoken-input scenarios, these concerns are further amplified \cite{lin2024listen, wu2025evaluating, arora2025landscape}. Spoken audio conveys rich paralinguistic information, including speaker gender, age, and accent, while also exhibiting substantial \emph{semantic} variation. Thus, observed performance disparities across subgroups may arise not only from biases inherited from the underlying LALMs, but also from \emph{confounding} factors such as semantic differences in spoken content and correlations among utterances produced by the same speaker. For example, if one demographic group is disproportionately associated with semantically more complex or information-dense utterances, a model may exhibit higher error rates for that group even in the absence of any intrinsic bias. If these factors are not properly controlled, fairness evaluations may overstate or misattribute disparities. 

Motivated by these challenges, this paper focuses on measuring audio understanding fairness in LALMs while explicitly controlling for semantic variation in spoken content. We propose a \emph{semantic-aware mixed-effects regression} method that incorporates sentence-level semantic embedding of \emph{reference} text as covariates while modeling speaker identity as a random effect. This approach enables more reliable assessment of subgroup performance gaps and helps disentangle the contributions of semantic content, speaker characteristics, and model behavior in observed disparities.

Existing work on statistical model-based fairness evaluation in audio understanding has largely focused on ASR tasks using traditional ASR systems \cite{liu2022model}. In these settings, semantic information is typically obtained from external representations, such as fastText \cite{bojanowski2017enriching} or BERT \cite{devlin2019bert}. In contrast, our work targets LALMs and considers a broader class of audio understanding tasks beyond ASR. Moreover, rather than relying on external semantic estimators, we propose a novel approach that leverages the \emph{same} LALM under evaluation to project reference text into a semantic vector space. By doing so, the semantic embeddings are directly aligned with the model's internal representation of the input, providing a more accurate and model-consistent control for semantic variation in fairness evaluation.

To the best of our knowledge, this work is the first to introduce a fairness evaluation framework for LALMs that explicitly accounts for semantic variation in spoken content. Our main contributions are threefold: (1) we propose a semantic-aware mixed-effects regression that includes semantic sentence embeddings as covariates while modeling speaker identity as a random effect, enabling more reliable fairness measurement in audio understanding tasks; (2) we develop new strategies that extract sentence embeddings of reference text from the same LALMs under evaluation; compared with external embedding estimators, the resulting embeddings are more closely aligned with the models' own perception and judgment of spoken content; and (3) we conduct extensive experiments on both ASR and audio question answering benchmarks, demonstrating that controlling for semantic and speaker-level confounding factors substantially reduces spurious fairness findings.

The rest of this paper is structured as follows. Section~\ref{related} discusses related work. Section \ref{method} introduces the semantic-aware mixed-effects regression method for fairness evaluations of LALMs in audio understanding tasks. Section~\ref{simulation} and Section~\ref{real} demonstrate the validity of the proposed approach on synthetic and real-world speech data, respectively. We conclude in Section \ref{conclusion}.

\section{Related Work}
\label{related}

\subsection{Fairness Evaluation in Audio Understanding}
Fairness in speech and audio systems has been extensively studied in the context of ASR, where performance disparities across demographic groups such as gender, age, and accent have been investigated \cite{tatman2017gender, koenecke2020racial}. Authors in \cite{liu2022model} propose a model-based statistical framework for ASR fairness evaluation that accounts for speaker-level dependencies using mixed-effects regression. Our work builds upon this framework and extends it in two key directions: (1) we target LALMs and consider a broader class of audio understanding tasks, beyond ASR with traditional hybrid or RNN-T speech models, and (2) we introduce strategies for extracting semantic representations from the same LALM under evaluation, providing model-consistent control for semantic variation rather than relying on external embeddings.

\subsection{Sentence Embeddings}
Extracting sentence representations has been a longstanding research topic \cite{bojanowski2017enriching, devlin2019bert, reimers2019sentence}. In contrast to these approaches, our work extracts sentence representations directly from the LALM under evaluation. This design choice ensures that the resulting embeddings reflect the model's own perception of semantic content, which is critical for controlling semantic confounding in fairness evaluation.

\section{Methods}
\label{method}

Suppose we aim to investigate fairness in audio understanding tasks for LALMs with respect to a factor variable of primary interest (e.g., speaker gender). For the $s$-th utterance in the evaluation dataset, we denote its factor level by $f(s)$ (e.g., male or female), where $f$ is a deterministic mapping. Our goal is to test whether the effect of this factor is \emph{statistically significant} on the measured task performance across its different levels.

We propose a semantic-aware mixed-effects regression and demonstrate how it can be utilized to quantify any performance gaps between subgroups in disparity and fairness analyses.

\subsection{Semantic-Aware Mixed-Effects Regression}
For illustration, we use the ASR task to demonstrate the proposed framework. After running speech recognition on an evaluation dataset using LALMs and obtaining the corresponding hypotheses, we compute utterance-level error statistics for subsequent analysis. Specifically, let $C^{\text{ins}}_{ij}$ and $C^{\text{del+sub}}_{ij}$ denote the number of insertion errors and the sum of deletion and substitution errors, respectively, for utterance $j$ from speaker $i$; let $N_{ij}$ be the number of words in the corresponding ground-truth transcription.

We model insertion errors using a \emph{Poisson regression}, and the sum of deletion and substitution errors using a \emph{Binomial regression}. This choice reflects the distinct nature of these error types: insertion errors correspond to count data without a fixed upper bound for a given utterance, making the Poisson distribution a natural modeling choice, whereas deletion and substitution errors are bounded by the number of reference words and can be viewed as Bernoulli trials at the word level, which are appropriately modeled using a Binomial distribution.

Let $x_{ij}$ denote the \emph{semantic sentence embedding} of  reference text for utterance $j$ from speaker $i$, which accounts for semantic variation across utterances in the evaluation set. Here, the reference text is defined as the concatenation of the ground-truth transcription of audio input and all accompanying textual content (e.g., instructions or  prompts) provided to the LALM under evaluation. We include $x_{ij}$ as a fixed-effect covariate in the model, and will discuss its computation in the next section.

In addition, we incorporate speaker-level random effects \cite{baltagi2008econometric, faraway2016extending} to capture speaker-specific characteristics. This modeling choice reflects the assumption that speakers are randomly sampled from a larger population and that our primary interest lies in estimating population-level effects rather than making inferences about individual speakers.

To measure the effect of factor $f(\cdot)$ on word-error-rate (WER) results across different subgroups, the proposed model is described as follows:
\begin{align}
r_i & \sim \mathcal{N}(0, \sigma^2), \\
C^{\text{ins}}_{ij} \mid \lambda_{ij} &\sim \text{Poisson}(\lambda_{ij}), \\
C^{\text{del+sub}}_{ij} \mid p_{ij} &\sim \text{Binomial}(N_{ij}, p_{ij}), \\
\log(\lambda_{ij}) &= \log(N_{ij}) + \mu^{\text{ins}}_{f(i)} + r_i + \theta^{\text{ins}\top} x_{ij}, \\
\text{logit}(p_{ij}) &= \mu^{\text{del+sub}}_{f(i)} + r_i + \theta^{\text{del+sub}\top} x_{ij},
\end{align}
where $r_i$ represents the speaker-level random effect that is independently sampled from a \emph{Gaussian} distribution with mean 0 and variance $\sigma^2$ which is learnable; $\lambda_{ij}$ denotes the expected number of insertion errors for utterance $j$ from speaker $i$; $p_{ij}$ denotes the per-word probability of a deletion or substitution error for the same utterance;
$\mu^{\text{ins}}_{f(i)}, \mu^{\text{del+sub}}_{f(i)}$ are fixed effects of the factor of interest $f(\cdot)$ for insertions and the sum of deletions and substitutions, respectively; $\theta^{\text{ins}}, \theta^{\text{del+sub}}$ are the corresponding regression coefficients for the semantic covariates.

The model can be fitted via \emph{maximum likelihood}, with the likelihood expressed as an integral over the random effects, which can be approximated using \emph{adaptive Gauss-Hermite quadrature} \cite{abramowitz1964handbook}. After fitting the model, we assess the effect of $f(\cdot)$ by leveraging the estimated fixed-effects coefficients and their variance-covariance matrices. Fairness disparities are then quantified by the ratio of the resulting expected WERs among different levels of $f(\cdot)$. Statistical inference on the WER ratios can be conducted by \emph{delta} method \cite{doob1935limiting} or \emph{bootstrap} \cite{efron1994introduction}.

Although we utilize ASR task as an example for illustration, the proposed framework is readily extensible to other audio understanding tasks for LALMs by adopting appropriate statistical distributional assumptions in the regression model for the task-specific response variable.

By including the semantic embeddings $x_{ij}$ as covariates, this framework explicitly controls for utterance-level semantic variation. This allows us to isolate the contribution of $f(\cdot)$ to audio understanding capabilities, providing a more precise and semantic-aware measure of fairness across subgroups. In practice, when the original embedding dimensionality is high, we optionally apply 
principal component analysis (PCA) to project the embeddings onto a lower-dimensional subspace that preserves the dominant semantic structure while ensuring stable and well-conditioned regression fitting.

\subsection{Extract Sentence Representations with LALMs}
We investigate two approaches for computing semantic embeddings of reference text using the same LALMs under evaluation. The resulting representations are incorporated as additional explanatory covariates in the regression models introduced in the previous section, enabling semantic-aware bias estimation. 

\subsubsection{Prompt LALMs with reference text}
For any reference text $\mathbf{x}=(x_{1}, \ldots, x_{n})$, passing it to the LALM under evaluation yields the hidden states
\begin{align}
\mathbf{h}_0, \mathbf{h}_1, \ldots, \mathbf{h}_L = \mathrm{LALM}(\mathbf{x}),
\end{align}
where $\mathbf{h}_0$ denotes the hidden state of the embedding layer, $\mathbf{h}_l$ for $l \in {1, \ldots, L}$ denotes the hidden state of the $l$-th Transformer \cite{vaswani2017attention} layer, and $L$ is the total number of Transformer layers. Let the final hidden state be written as $\mathbf{h}_L = (h_{L1}, \ldots, h_{Ln'})$, where $n'$ is the number of tokens after tokenization.

The hidden state of the final token, $h_{Ln'}$, can be used as the sentence embedding of $\mathbf{x}$. Alternatively, other aggregation or pooling strategies over the hidden states can be employed. In this work, we consider an aggregation method that averages the hidden states from the first, middle, and final Transformer layers across all tokens
\begin{align}
\frac{1}{3}\left(
\frac{1}{n'} \sum_{k=1}^{n'} h_{1k}
+
\frac{1}{n'} \sum_{k=1}^{n'} h_{\lfloor L/2 \rfloor, k}
+
\frac{1}{n'} \sum_{k=1}^{n'} h_{Lk}
\right).
\end{align}
This aggregation is motivated by the observation that earlier Transformer layers tend to encode lower-level lexical features, while deeper layers progressively capture higher-level semantic abstractions \cite{zhao2023survey}. By combining representations across different depths, the resulting embedding provides a more comprehensive summary of the input.

We refer to this method as {``Embed-AGG''}.

\subsubsection{Prompt LALMs with explicit one word limitation}

The prompt template with an explicit one-word limitation (EOWL) was first introduced in \cite{jiang2023scaling}. It provides a simple and direct way to instruct LLMs to summarize the meaning of a sentence using a single word. Under this template, for any reference text $\mathbf{x}$, we query the LALM under evaluation as
\begin{align}
\text{LALM}({\emph{\text{This sentence: }}``\mathbf{x}"\emph{\text{ means in one word:}}}).
\end{align}
In contrast to \cite{jiang2023scaling}, which leverages hidden states, we use the logits of the LALM's next-token prediction as the representation of $\mathbf{x}$. The motivation is that the EOWL prompt encourages the model to condense the semantic content of the entire sentence into a single-word prediction. Thus, the associated output logits can be viewed as a compact semantic representation of the sentence. We refer to this method as {``Embed-EOWL''}.

Note that extracting semantic embeddings from the same LALM under evaluation is a deliberate design choice. The goal of the semantic covariate is to control for variation in how the model perceives input difficulty; using the model's own representations provides the most faithful characterization of this variation.

\section{Simulation Experiments}
\label{simulation}

Under the scenario where the factor of interest (e.g., speaker gender) has no true effect on the response quality of LALMs, substantial differences in the semantic content of the audio inputs across factor levels can nevertheless lead to spurious findings of unfairness if not properly accounted for. In this section, we perform simulation experiments to demonstrate that the proposed semantic-aware regression framework effectively addresses semantic confounding in fairness measurements for LALMs, thereby reducing bias in effect estimation and mitigating false positive conclusions of unfairness.

\subsection{Setups}
Suppose we want to study the effect of speaker gender on the response accuracy in audio question answering task for LALMs.

First, we prompt the Llama-3 70B model \cite{grattafiori2024llama} to generate 2{,}000 text questions. Using different prompt instructions, half of the questions are designed to be short, clear, and primarily based on common-sense knowledge, while the other half are longer, more detailed, and require deeper reasoning that draws on knowledge from multiple domains. We refer to these two subsets as the \emph{simple} question set and the \emph{hard} question set, respectively.

We then apply a text-to-speech (TTS) system to convert the text questions into audio questions. Specifically, we randomly select 200 questions from the simple set and 800 questions from the hard set to be synthesized using \emph{female} speakers, and 800 questions from the simple set and 200 questions from the hard set to be synthesized using \emph{male} speakers. We use the Qwen2-Audio model \cite{chu2024qwen2} as the LALM to generate textual responses to these audio questions.

In this simulation, assume that the response accuracy scores follow a $\mathcal{N}(5,1)$ distribution for questions from the hard set and a $\mathcal{N}(6,1)$ distribution for questions from the simple set. Therefore, under this data-generating process, speaker gender has no effect on response accuracy. This design intentionally induces correlation between gender and semantic difficulty while ensuring that gender has no causal effect on response accuracy.

\subsection{Methods and Results}
As a vanilla approach, we compute the average response scores for the male and female groups and report their ratio, with 95\% confidence intervals estimated via bootstrap. For the semantic-aware regression approach, we extract sentence embeddings using the same Qwen2-Audio model with both the {Embed-AGG} and {Embed-EOWL} methods. PCA is then applied to reduce the embedding dimensionality to 8, which retains the majority (around 83\%) of the total variance and yield stable regression estimates in practice. We subsequently fit a linear regression model that includes the reduced semantic representations as covariates and estimate the ratio of response scores between the male and female groups, as well as the corresponding confidence intervals. For comparison, we also include a regression method that excludes semantic information although it is equivalent to the vanilla estimation.

\begin{table}[ht]
 \caption{Comparison of vanilla and semantic-aware methods for gender fairness evaluation on simulated data.}
  \centering
  \resizebox{\columnwidth}{!}{%
  \begin{tabular}{l|c|c|c}
    \toprule
    & \multicolumn{3}{|c}{\bf{\texttt{Simulated Data}}} \\
    \cmidrule(r){2-4}       
    & \multicolumn{3}{|c}{\emph{Male-Female Score Comparison}} \\
    \cmidrule(r){2-4}    
    \bf{Method}& \shortstack{\emph{Ratio}} & \shortstack{\emph{Confidence} \\ \emph{Interval}} & \shortstack{\emph{Is False} \\ \emph{Positive?}} \\
    \midrule
    {Vanilla Estimation} & 1.124 & (1.103, 1.144) & Y\\
    {Regression w/o Semantics} & 1.124 & (1.104, 1.144) & Y\\    
    {Semantic-Aware: {Embed-AGG}} & 1.003 & (0.983, 1.024) & N \\
    {Semantic-Aware: {Embed-EOWL}} & 1.000 & (0.980, 1.021) & N \\
    \bottomrule
  \end{tabular}
  }
  \label{tab:sim}
\end{table}

Table~\ref{tab:sim} summarizes the estimated male-female response accuracy score ratios obtained using different evaluation methods. Both the vanilla estimation approach and the regression model without semantic covariates yield identical ratio estimates (1.124) with confidence intervals that exclude 1, leading to false positive conclusions of gender unfairness despite the ground truth being gender-neutral. In contrast, the proposed semantic-aware regression methods produce ratio estimates close to 1 with confidence intervals covering 1, correctly indicating no statistically significant performance gap between male and female speakers. These results demonstrate that failing to control for semantic variation can induce spurious fairness violations, while incorporating semantic sentence embeddings effectively mitigates semantic confounding and yields more reliable fairness assessments.

\begin{figure}[htbp]
    \centering
    \includegraphics[width=0.35\textwidth]{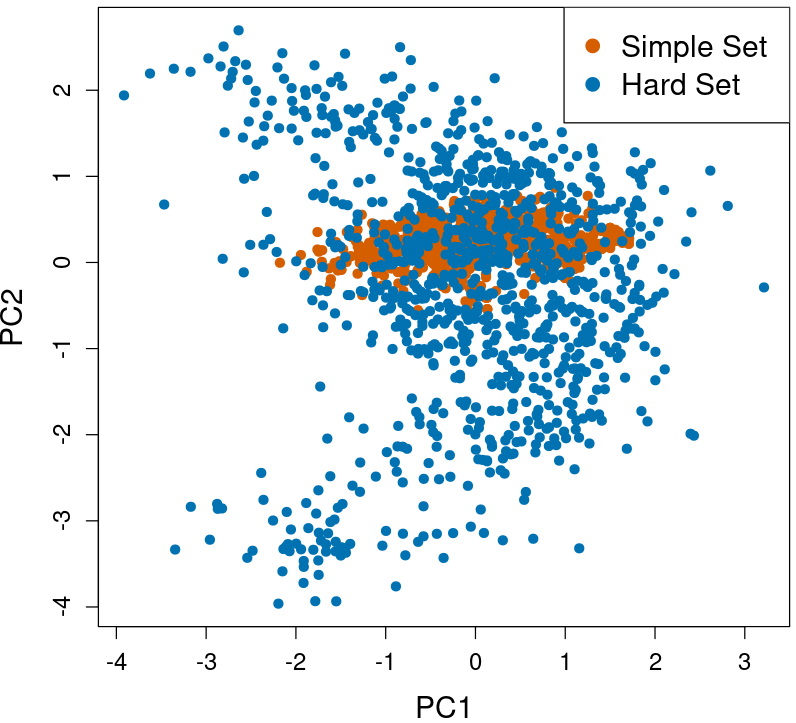}
    \caption{PCA visualization of sentence embeddings.}
    \label{fig:pca}
\end{figure}

After applying PCA to further project the 8-dimensional Embed-AGG into 2 dimensions, Figure~\ref{fig:pca} illustrates that the embeddings extracted from Qwen2-Audio clearly differentiate between simple and hard questions. 
This separation indicates that the extracted sentence embeddings successfully capture the underlying semantic difficulty of the questions.

\section{Real Data Experiments}
\label{real}

In this section, we apply the proposed semantic-aware mixed-effects regression to real speech data for fairness analysis.

\subsection{Setups}
We consider the following two evaluation datasets in this study:
\begin{itemize}
\item \texttt{LibriSpeech}~\cite{panayotov2015librispeech}. It is a widely used ASR benchmark. We use the evaluation splits, including \texttt{Test-Clean}, which contains 2{,}620 utterances from 40 speakers (20 male), and \texttt{Test-Other}, which contains 2{,}939 utterances from 33 speakers (16 male). We evaluate ASR fairness on gender;
\item \texttt{AIR-Bench-Chat}~\cite{yang2024airbench}. It is a benchmark of audio understanding questions designed to evaluate the ability of LALMs to comprehend complex audio inputs and follow human instructions. In this experiment, we select the subset with available gender annotations, consisting of 436 utterances from male speakers and 163 utterances from female speakers. For evaluation, we use Llama-3 70B model as a judge to score the generated answers from LALMs. The scores range from 1 to 10 and are based on usefulness, relevance, accuracy, and comprehensiveness. We then analyze these scores to assess gender fairness. Since the LLM judge evaluates only textual responses without access to speaker information, its scoring is applied uniformly across subgroups and is unlikely to confound gender comparisons.

\end{itemize}
For both datasets, we use Qwen2-Audio as the LALM to generate textual responses, either by transcribing audio inputs or by answering audio-based questions.

\subsection{Methods and Results}
Table~\ref{tab:real1} presents the aggregated performance metrics for male and female speakers across the evaluated datasets. For the ASR task on \texttt{LibriSpeech} datasets, we present the average WER, while for \texttt{AIR-Bench-Chat} we report the average response accuracy score. The corresponding male-female ratios provide a naive measure of group-level performance differences. While such aggregated statistics are intuitive and commonly adopted, they do not account for potential confounding factors such as speaker-specific characteristics or semantic variation.

\begin{table}[ht]
  \centering
  \caption{Aggregated male and female performance comparison across evaluation datasets.}
  
  \label{tab:real1}
  \resizebox{\columnwidth}{!}{%
  \begin{tabular}{l|cc|c}
    \toprule
    & \multicolumn{3}{c}{\emph{Aggregated WER or Score}} \\
    \cmidrule(r){2-4}
    & \emph{Male} & \emph{Female} & \emph{Ratio} \\
    \midrule
    {\texttt{LibriSpeech Test-Clean}} (WER) & 3.21 & 4.47 & 0.719 \\ 
    {\texttt{LibriSpeech Test-Other}} (WER) & 6.89 & 5.93 & 1.162 \\
    {\texttt{AIR-Bench-Chat}} (Score) & 6.02 & 6.21 & 0.969 \\
    \bottomrule
  \end{tabular}
  }
\end{table}

We examine regression-based models that progressively incorporate additional structure, including fixed-effect regression without speaker- or semantic-level controls, mixed-effects regression with speaker-level random effects (when speaker identifiers are available), and our proposed semantic-aware mixed-effects regression models that further include the semantic embeddings of references as covariates. For the semantic-aware models, we extract sentence embeddings using Qwen2-Audio with both the {Embed-AGG} and {Embed-EOWL} methods. PCA is then applied to reduce the embedding dimensionality to 8, which retains the majority of the total variance while ensuring stable and well-conditioned regression estimation. This systematic comparison allows us to isolate the impact of speaker-level dependencies and semantic confounding on fairness measurements. 

Tables~\ref{tab:real2} and \ref{tab:real3} show the comparison of different methods for gender fairness evaluation on \texttt{LibriSpeech} data. On the \texttt{Test-Clean} and \texttt{Test-Other} sets, vanilla estimation and regression models without speaker or semantic controls indicate significant male-female WER gaps. In contrast, models that account for speaker-level random effects eliminate much of this bias, and the proposed semantic-aware mixed-effects regression further attenuates the estimated gaps, rendering them statistically insignificant. This suggests that previously observed disparities are largely driven by speaker-specific characteristics and sentence-level linguistic differences rather than gender itself.

\begin{table}[ht]
 \caption{Comparison of different methods for gender fairness evaluation on LibriSpeech Test-Clean data.}
  \centering
  \resizebox{\columnwidth}{!}{%
  \begin{tabular}{l|c|c|c}
    \toprule
    & \multicolumn{3}{|c}{{\texttt{LibriSpeech Test-Clean}}} \\
    \cmidrule(r){2-4}
    & \multicolumn{3}{|c}{\emph{Male-Female WER Comparison}} \\
    \cmidrule(r){2-4}    
    \bf{Method}& \shortstack{\emph{Ratio}} & \shortstack{\emph{Confidence} \\ \emph{Interval}} & \shortstack{\emph{Is Stat.} \\ \emph{Significant?}} \\
    \midrule
    {Vanilla Estimation} & 0.719 & (0.553, 0.938) & Y\\
    {Regression w/o Spk. w/o Sem.} & 0.719 & (0.658, 0.786) & Y\\  
    {Regression w/ Spk. w/o Sem.} & 0.777 & (0.600, 1.011) & N\\  
    {Spk. Sem.-Aware: {Embed-AGG}} & 0.802 & (0.619, 1.042) & N \\
    {Spk. Sem.-Aware: {Embed-EOWL}} & 0.822 & (0.634, 1.073) & N \\
    \bottomrule
  \end{tabular}
  }
  \label{tab:real2}
\end{table}

\begin{table}[ht]
  \caption{Comparison of different methods for gender fairness evaluation on LibriSpeech Test-Other data.}
  \centering
  \resizebox{\columnwidth}{!}{%
  \begin{tabular}{l|c|c|c}
    \toprule
    & \multicolumn{3}{|c}{{\texttt{LibriSpeech Test-Other}}} \\
    \cmidrule(r){2-4}
    & \multicolumn{3}{|c}{\emph{Male-Female WER Comparison}} \\
    \cmidrule(r){2-4}    
    \bf{Method}& \shortstack{\emph{Ratio}} & \shortstack{\emph{Confidence} \\ \emph{Interval}} & \shortstack{\emph{Is Stat.} \\ \emph{Significant?}} \\
    \midrule
    {Vanilla Estimation} & 1.162 & (1.013, 1.329) & Y\\
    {Regression w/o Spk. w/o Sem.} & 1.162 & (1.088, 1.244) & Y\\  
    {Regression w/ Spk. w/o Sem.} & 1.125 & (0.875, 1.450) & N\\  
    {Spk. Sem.-Aware: {Embed-AGG}} & 1.020 & (0.811, 1.289) & N \\
    {Spk. Sem.-Aware: {Embed-EOWL}} & 1.017 & (0.801, 1.294) & N \\
    \bottomrule
  \end{tabular}
  }
  \label{tab:real3}
\end{table}

Table~\ref{tab:real4} shows the comparison of different methods for gender fairness evaluation on the \texttt{AIR-Bench-Chat} dataset. All methods show no statistically significant gender differences in response quality. Notably, the semantic-aware regression yields ratios closest to unity, confirming that controlling for semantic content leads to more stable and interpretable fairness assessments in audio understanding tasks. Note that speaker-level random effects cannot be applied to this dataset, as speaker identifiers are not provided in the original benchmark.

\begin{table}[ht]
 \caption{Comparison of different methods for gender fairness evaluation on AIR-Bench-Chat data.}
  \centering
  \resizebox{\columnwidth}{!}{%
  \begin{tabular}{l|c|c|c}
    \toprule
    & \multicolumn{3}{|c}{{\texttt{AIR-Bench-Chat}}} \\
    \cmidrule(r){2-4}
    & \multicolumn{3}{|c}{{Male-Female Score Comparison}} \\
    \cmidrule(r){2-4}    
    \bf{Method}& \shortstack{\emph{Ratio}} & \shortstack{\emph{Confidence} \\ \emph{Interval}} & \shortstack{\emph{Is Stat.} \\ \emph{Significant?}} \\
    \midrule
    {Vanilla Estimation} & 0.969 & (0.893, 1.055) & N\\
    {Regression w/o Semantics} & 0.969 & (0.891, 1.055) & N\\    
    {Semantic-Aware: {Embed-AGG}} & 1.004 & (0.920, 1.101) & N \\
    {Semantic-Aware: {Embed-EOWL}} & 1.002 & (0.924, 1.093) & N \\
    \bottomrule
  \end{tabular}
  }
  \label{tab:real4}
\end{table}

The Embed-AGG and Embed-EOWL methods perform similarly in practice; however, Embed-AGG is more effective, as its lower-dimensional PCA projection explains a larger proportion of the variance in the original embedding space.

Overall, these results demonstrate that failing to control for semantic and speaker-level confounding factors can lead to false conclusions of unfairness. The proposed semantic-aware mixed-effects framework provides a more reliable and principled approach for measuring fairness in LALMs across audio understanding tasks.

While our experiments focus on Qwen2-Audio and gender as the primary attribute, we emphasize that the proposed framework is model-agnostic and attribute-agnostic by design. It is natural to extend the evaluation to additional models and demographic attributes.

\section{Conclusions}
\label{conclusion}
In this work, we present a semantic-aware mixed-effects regression model for measuring fairness in LALMs. By controlling for sentence-level semantic variation, the proposed approach provides more reliable estimates of subgroup performance gaps than conventional methods. Experimental results show that many apparent disparities can be attributed to semantic or speaker confounding rather than intrinsic model bias. This paper highlights the importance of semantic control in fairness evaluations and offers a general framework applicable to a wide range of audio understanding tasks.

\bibliographystyle{IEEEtran}
\bibliography{references}

\end{document}